\documentclass{article}

\usepackage{microtype}
\usepackage{graphicx}
\usepackage{subfigure}
\usepackage{booktabs} % for professional tables

\usepackage{hyperref}

\usepackage[accepted]{icml2024}

\usepackage{amsmath}
\usepackage{amssymb}
\usepackage{mathtools}
\usepackage{amsthm}

\usepackage[capitalize,noabbrev]{cleveref}

\theoremstyle{plain}

\theoremstyle{definition}

\theoremstyle{remark}

\usepackage[textsize=tiny]{todonotes}
\usepackage{enumitem}
\usepackage{bm}
\usepackage{subcaption}

\usepackage{xspace}
\newcommand{\method}{RoboDreamer\xspace}

\definecolor{MyDarkBlue}{rgb}{0,0.08,1}
\definecolor{MyDarkGreen}{rgb}{0.02,0.6,0.02}
\definecolor{MyDarkRed}{rgb}{0.8,0.02,0.02}
\definecolor{MyDarkOrange}{rgb}{0.40,0.2,0.02}
\definecolor{MyPurple}{RGB}{111,0,255}
\definecolor{MyRed}{rgb}{1.0,0.0,0.0}
\definecolor{MyGold}{rgb}{0.75,0.6,0.12}
\definecolor{MyDarkgray}{rgb}{0.66, 0.66, 0.66}

\def\Gcal{\mathcal{G}}
\def\Xcal{\mathcal{X}}
\def\Ccal{\mathcal{C}}
\def\Ncal{\mathcal{N}}
\def\Acal{\mathcal{A}}

\begin{document}
\twocolumn[
\icmltitle{ RoboDreamer:  Learning Compositional  World Models 
for Robot Imagination}
% with Multimodal Instructions


\icmlsetsymbol{equal}{*}

\begin{icmlauthorlist}
\icmlauthor{Siyuan Zhou}{hkust}
\icmlauthor{Yilun Du}{mit}
\icmlauthor{Jiaben Chen}{ucsd}
\icmlauthor{Yandong Li}{google}
\icmlauthor{Dit-Yan Yeung}{hkust}
\icmlauthor{Chuang Gan}{umass,mit-ibm}
\end{icmlauthorlist}

\icmlaffiliation{hkust}{Hong Kong University of Science and Technology}
\icmlaffiliation{mit}{Massachusetts Institute of Technology}
\icmlaffiliation{ucsd}{University of California, San Diego}
\icmlaffiliation{google}{Google Research}
\icmlaffiliation{umass}{University of Massachusetts Amherst}
\icmlaffiliation{mit-ibm}{MIT-IBM Watson AI Lab}

 \begin{center}{
        \hypersetup{urlcolor=red}
        \url{https://robovideo.github.io/}
    }\end{center}


\icmlkeywords{Machine Learning, ICML}

\vskip 0.3in
]



\printAffiliationsAndNoticeArxiv{}  % leave blank if no need to mention equal contribution

\begin{abstract}
Text-to-video models have demonstrated substantial potential in robotic decision-making, enabling the imagination of realistic plans of future actions as well as accurate environment simulation. However, one major issue in such models is generalization -- models are limited to synthesizing videos subject to language instructions similar to those seen at training time. This is heavily limiting in decision-making, where we seek a powerful world model to synthesize plans of unseen combinations of objects and actions in order to solve previously unseen tasks in new environments. To resolve this issue, we introduce RoboDreamer, an innovative approach for learning a compositional world model by factorizing the video generation.  We leverage the natural compositionality of language to parse instructions into a set of lower-level primitives, which we condition a set of models on to generate videos. We illustrate how this factorization naturally enables compositional generalization,  by allowing us to formulate a new natural language instruction as a combination of previously seen components. We further show how such a factorization enables us to add additional multimodal goals, allowing us to specify a video we wish to generate given both natural language instructions and a goal image.  Our approach can successfully synthesize video plans on unseen goals in the RT-X, enables successful robot execution in simulation, and substantially outperforms monolithic baseline approaches to video generation.

% However, existing models primarily rely on language guidance, overlooking the importance of incorporating other modalities crucial for informed decision-making. In light of this, our objective is to develop video generation models that can be conditioned on diverse inputs, ranging from text and images to sketches. These models should seamlessly integrate information from multiple modalities to enhance their generation capabilities in decision-making scenarios. In pursuit of this goal, we introduce RoboDreamer, a versatile video model designed to approach the challenges posed by general-purpose world models in decision-making tasks. Notably, RoboDreamer achieves the compositional generation ability by employing a self-learning strategy and factorizing different modalities within its diffusion models.
\end{abstract}
\section{Introduction}

\begin{figure}[!t]
    \centering
    \includegraphics[width=\linewidth]{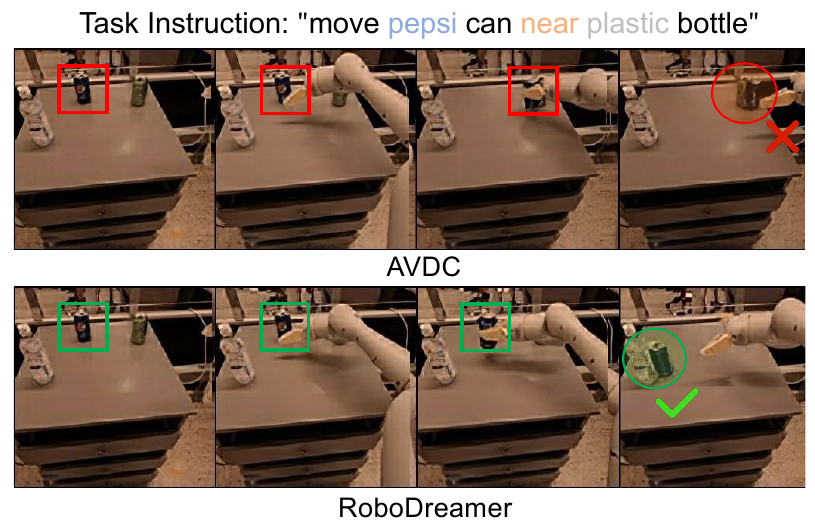}
   
    \caption{\textbf{Compositional Action Specification.} When existing text-to-video models (AVDC~\cite{ko2023learning}) are given unusual combinations of language instructions, they are unable to synthesize videos that align accurately with these descriptions. \method factorizes the generation compositionally, enabling generalization to novel combinations of language. 
    }
    \label{fig:fail_succ}
    \vspace{-15pt}
\end{figure}

% Text-to-video models have recently demonstrated significant potential in decision-making, laying a solid foundation for a variety of policies, dynamics models and planners that are built from such models.

% Text-to-video models have recently emerged as powerful tools for decision-making, laying a solid foundation for the development of policies, dynamic models, and planners~\cite{du2023learning, ajay2023compositional, yang2023learning}. These models have demonstrated their capability to perform planning by synthesizing detailed future frames in videos. They efficiently utilize a vast array of video data from diverse sources including the internet, human demonstrations~\cite{Damen2018EPICKITCHENS, grauman2022ego4d}, as well as from the real world, where the collection of corresponding actions poses a significant challenge.
% Furthermore, text-to-video models could handle datasets with large diversity across different robots, environments, and tasks, even though the unification and alignment of state and action spaces on different setups prove to be more costly and complex~\cite{padalkar2023open}.

\begin{figure*}[t]
    \centering
    \includegraphics[width=1.0\linewidth]{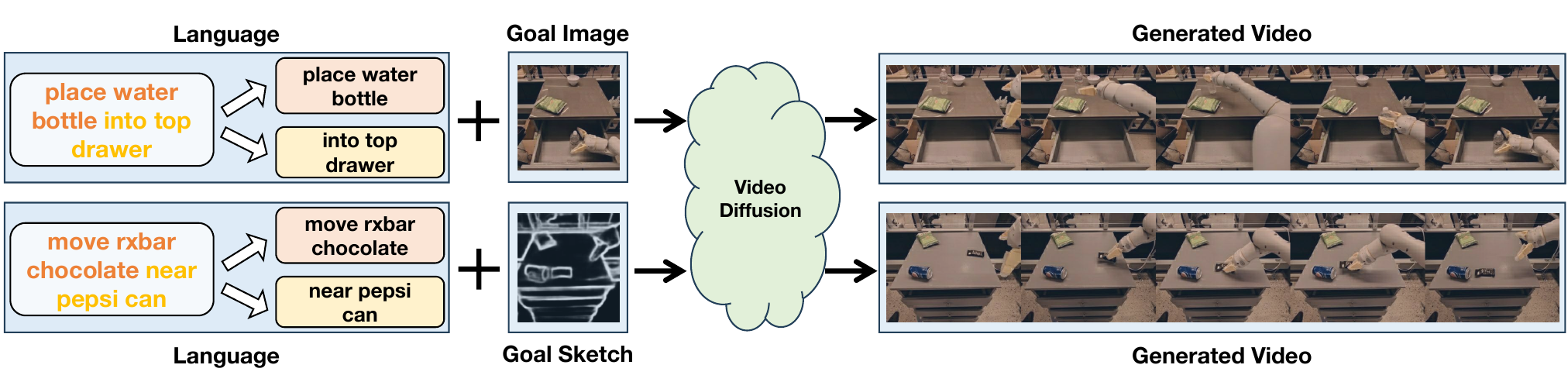}
    \vspace{-10pt}
    \caption{\textbf{Compositional World Models.} Given language instructions and multimodal instructions such as goal images and sketches, our approach factorizes the generation into a composition of diffusion models conditioned on inferred components. This enables our approach to generalize to both new combinations of language and multimodal input. }
    \label{fig:comp_wm}
    \vspace{-10pt}
\end{figure*}

 Text-to-video models\cite{ho2022imagen, singer2022make, villegas2022phenaki} have seen extensive development in the setting of AI content generation, where models can generate high-quality videos given short text descriptions of motions. 
Such models have recently been applied in robotics, demonstrating significant potential in the development of policies, dynamic models, and planners~\cite{du2023learning, ajay2023compositional, yang2023learning}.
% However, the distribution of both videos and language in robotics differs substantially from content generation. 
% However, when applied to robotics, these models 
However, while natural language commands found in content generation typically focus on the global motion of a scene, natural language actions in robotics revolve around precise spatial rearrangements between objects.

% instructions, particularly in capturing spatial reasoning and object relations critical for task execution. 

Such commands, such as ``move pepsi can near plastic bottle." remain challenging for existing models. As shown in Fig. \ref{fig:fail_succ}, existing methods generate a video where pepsi can is placed near green can, failing to accurately capture the specified object relationship. Furthermore, these challenges become even more pronounced in scenarios where language instructions deviate from those encountered during training time, especially in reinforcement learning datasets where the data are scarce and natural language instructions are highly biased.

In response, we introduce \method, a compositional world model capable of \textit{factorizing the video generation process} by leveraging the \textit{inherent compositionality of natural language}. This model is designed to equip conventional text-to-video generation systems with the ability to perform compositional reasoning. By utilizing a text parser, we dissect language instructions into a set of primitives, isolating actions and the spatial relationships between objects. These parsed components then serve as distinct conditions for a compositional set of diffusion models, enhancing the ability of each model to capture the nuances of spatial relationships among objects. Additionally, by decomposing a text goal into a set of text components, our approach naturally generalizes to new combinations of language as long as each parsed component is in distribution. This is crucially important in robotics, where there is a lack of systematic data covering all possible actions in an environment and a need to be able to generalize to new unseen actions.

% On the other hand, this discrepancy highlights the models' difficulty in precisely interpreting and executing tasks that require nuanced understanding of spatial relationships and object interactions based on textual instructions alone. In contrast to the limitations of purely text-based task instructions, humans demonstrate a versatile ability to depict the tasks in terms of the desired goal states through imagination, represented in multiple modalities beyond language, and guide their own courses of goal-directed plans and actions from the imagination ~\cite{SpaldingThePO}. 

While natural language is one representation to specify tasks a robot should accomplish, it is high-level and abstract, making it difficult to convey the precise nuances of motion over target goal configurations.  Other modalities of information, such as a goal image provide much more detailed information on the final goal we wish to achieve. We illustrate how \method also enables us to compose across \textit{multimodal specifications} to flexibly specify goals at inference time. In contrast to prior work leveraging video models for robotics, we further condition video generation on multimodal task specification goal images and goal sketches. These modalities, particularly rich in spatial information, play a crucial role in clarifying ambiguities inherent in task execution instructions. Goal sketches, in particular, offer an intuitive and user-friendly means for spontaneous and on-the-fly task expression, akin to language instructions.

To compose these multimodal specifications together in \method, we similarly factorize generation into a set of models jointly conditioned on language components as well as other multimodal components (Figure~\ref{fig:comp_wm}). We illustrate how this approach enables us to richly specify and generate videos given a large set of specified conditions, enabling us at inference time to combine larger and new combinations of both language and multimodal specifications, even if such paired conditions are not available at training time. Prior approaches, such as  ControlNet~\cite{zhang2023adding} introduce an additional encoder upon pre-trained text-to-image models to tackle this challenge, but this requires the availability of paired data across language and multimodal inputs and is limited at inference time to a similar combination of inputs that are seen at training time.

Our contributions are three-fold. 
\begin{itemize} 
\item We introduce \method, a compositional world model capable of factorizing the video generation process by leveraging the inherent compositionality of natural language.
\item We illustrate how \method also allows us to combine multimodal information, enabling us to combine goal information from images with natural language.
\item We empirically demonstrate that \method achieves strong alignment with tasks under multi-modal instructions and promising results when deploying on robot manipulation tasks.
\end{itemize}

\begin{figure*}[t]
    \centering
    \includegraphics[width=1.0\linewidth]{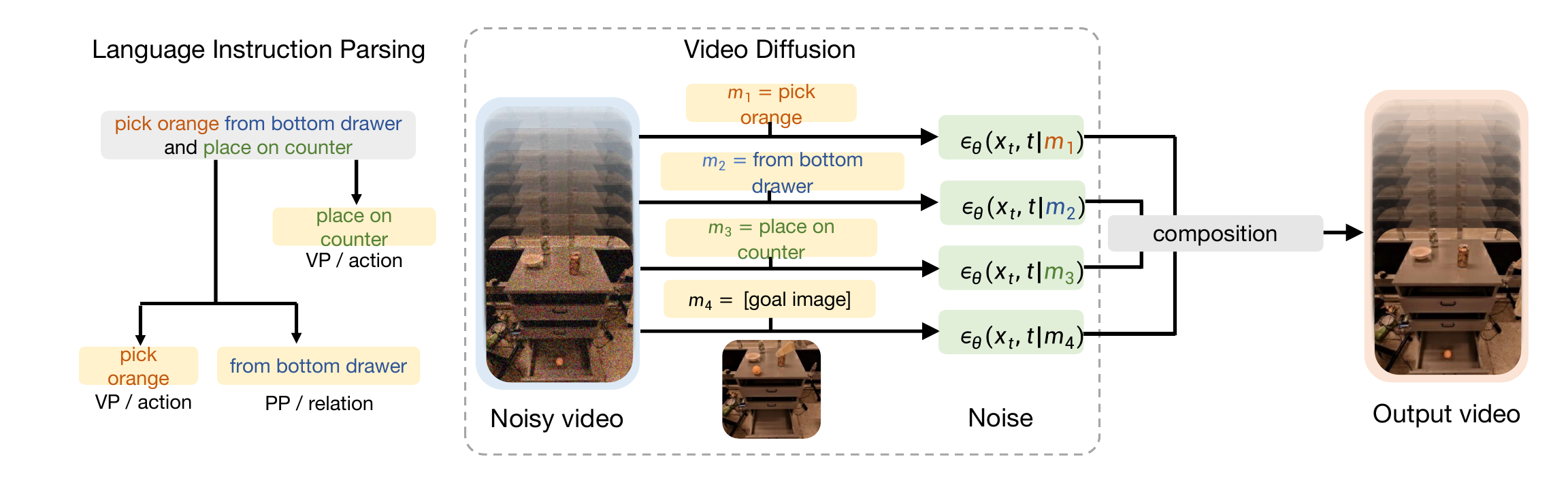}
    \vspace{-18pt}
    \caption{\textbf{Overall framework of \method.} On the left, We leverage the natural compositionally of language to parse instructions into components like action phrases and relation phrases. On the right, we show how \method composes multiple components. 
    }
    \vspace{-10pt}
    \label{fig:method}
\end{figure*}
\section{Background}

We first provide background information on how text-conditioned video generation can serve as a world model for robotics. As an initial step, we introduce formalism on how text-conditioned video generation can be used for planning and how we can implement such video level planning in Section~\ref{sect:planning}. We then discuss how to execute video plans in Section~\ref{sect:video_execute}
and subsequently use videos in a close-loop manner for task completion. 

\subsection{Planning with Text-Conditioned Video Generation}
\label{sect:planning}

We formulate task planning as a text-conditioned video generation problem using the Unified Predictive Decision Process (UPDP) abstraction from ~\citep{du2023learning}. 
Formally, a UPDP is defined as a tuple $\Gcal = \langle \Xcal, \Ccal, H, \rho\rangle$,  where $\Xcal$ denotes the observation space of images, $\Ccal$ denotes the space of textual task descriptions, $H\in \Ncal$ is a finite horizon length, and $\rho(\cdot| x_0, c): \Xcal \times\Ccal\rightarrow \Delta(\Xcal^H)$ as a conditional video generator which synthesizes a video given a text description $c$ and starting observation $x_0$. Given a UPDP $\Gcal$, 
we then use a trajectory-task conditioned policy $\pi(\cdot| \{x_h\}_{h=0}^H, c): \Xcal^{H+1}\times\Ccal\rightarrow \Delta(\Acal^{H})$ to infer executable actions from synthesized videos.

Under this decision process, decision-making simply corresponds to learning $\rho$, which synthesizes videos of future image states given a natural language instruction $c$. This enables us to convert planning directly into a text-to-video generation problem. To implement this generation problem, we use the video diffusion model and use the base source code from ~\citep{ko2023learning}.

% To implement video generation, we use a text-to-video diffusion model is used to 

\subsection{Executing Videos Plans}
\label{sect:video_execute}

Given a synthesized video plan  $\tau = [x_1, \ldots, x_H]$, we follow~\citep{du2023learning} and use an inverse dynamics model $\pi(\cdot)$ to infer actions to execute to realize the video plan. The policy takes as input two adjacent image observations $x_{t}$ and $x_{t+1}$ in the synthesized video $\tau$ and outputs an action $a$ to execute. We sequentially execute inferred actions starting from $x_1$ to $x_{H-1}$ in the video. 

To account for intermediate estimation error from predicting actions using an inverse dynamics model, we predict videos in a close-loop manner, where we periodically regenerate new video plans and execute actions based on this new plan.

% several actions from the inverse dynamics model, a new synthesized video plan is subsequently generated.

%\input{tables/parse_example}

\begin{figure*}[t]
    \centering
    \includegraphics[width=1.0\linewidth]{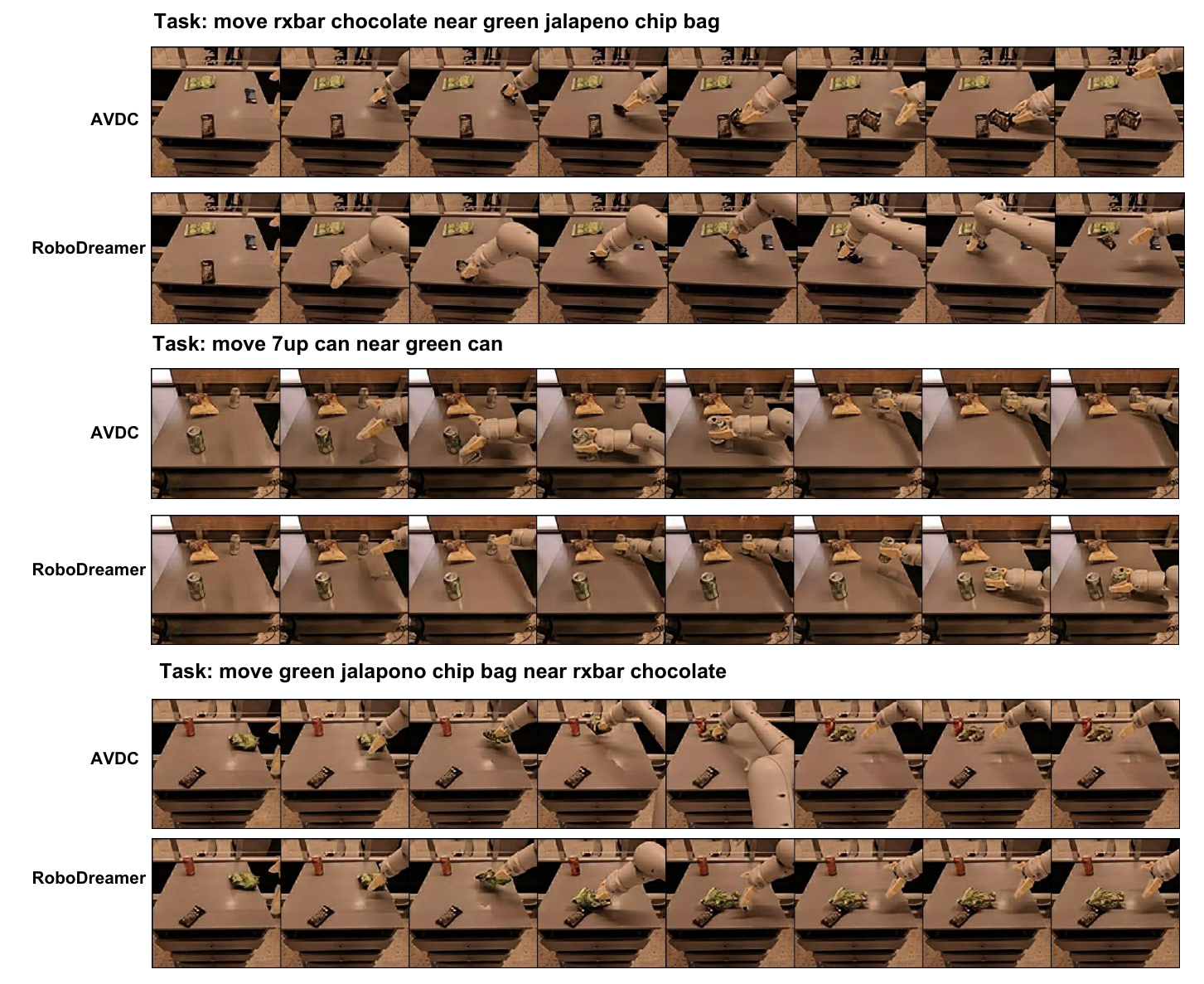}
    % \includegraphics[width=1.0\linewidth]{figures/comp/demo1.pdf}
    % \vspace{-10pt}
    % \includegraphics[width=1.0\linewidth]{figures/comp/demo2.pdf}
    % \includegraphics[width=1.0\linewidth]{figures/comp/demo3.pdf}
    \vspace{-3pt}
    \caption{\textbf{Zero-Shot Video Generation.} Given novel combinations of natural language, \method is able to substantially more accurately synthesize videos than a single monolithic text-to-video model.}
    \label{fig:zero-shot}
    \vspace{-10pt}
\end{figure*}

\section{RoboDreamer}

In this section, we describe our proposed \method in detail.
% We first present how \method generates videos based on multi-modal instructions.
To construct a compositional world model, \method first uses a text parser to convert the compositional structure of language into a set of shared components in Section~\ref{sect:text_parse}. Given these shared components, we illustrate how we can use compositional generation to guarantee generalization to novel combinations of these components in Section~\ref{sect:comp}. Finally, we illustrate how such compositional components can be extended to multimodal conditions such as images or sketches for detailed compositional specification of synthesized plans in Section~\ref{sect:multimodal}.

\subsection{Text Parser}
\label{sect:text_parse}

In contrast to many existing applications of text-to-video models in AI content creation, in the robotics setting, we are interested in synthesizing accurate video plans given detailed natural language instructions of actions. To solve robotics tasks, it is important that video models can synthesize actions that precisely rearrange one object with detailed specified relations with respect to nearby objects, including rearrangements not seen at training. However, reasoning and correctly generating videos subject to such object spatial relations is often challenging for existing models, especially on unseen textual descriptions.

To construct models that can more accurately synthesize spatial relations, we propose to decompose each spatial relation phrase into a set of compositional components. In particular, the actions of the tasks usually correspond to the verb phrase in the language and the object spatial relations correspond to the prepositional phrase after the verb phrase. Thus, given a text action instruction $L$, we decompose the instruction into a set of verb and propositional phrases $l_i$ which we use to condition a set of diffusion models on.

Take the task \textit{place water bottle into bottom drawer
} as an example. From this sentence, we parse the verb phrase \textit{place water bottle} as the actions of the task and the prepositional phrase \textit{into bottom drawer} as the object spatial relations. We utilize the pre-trained parser~\cite{kitaev2018multilingual} and the rule-based approach to parse language instructions based on such characteristics. We provide a schematic of the text parsing in Figure~\ref{fig:method}.

\subsection{Compositional Generation}
\label{sect:comp}

Given a natural language instruction $L$ that is parsed into a set of language components $\{l_i \}_{i=1:N}$, we propose to formulate our text-to-video model generative model $p_\theta(\tau|L)$ as a product of individual generative models defined on each parses language subcomponent $l_i$
\begin{equation}
    \label{eqn:product}
    p_\theta(\tau|L) \propto \prod_{i=1:N} p_\theta(\tau|l_i)^{\frac{1}{N}}.
\end{equation}
Note that Eqn~\ref{eqn:product} naturally enables {\it compositional generalization} -- given unseen combinations of natural language instructions $L$, our probabilistic expression in Eqn~\ref{eqn:product} will generalize perfectly as long as each parsed components $l_i$ are in distribution. Thus, we can naturally map the compositionally of language into the video space through syntactic parsing.
% ~\sy{We are not always parsing the language instructions. For example A->B+C, we may train on A, or (B, C) or B or C}

To train our probabilistic expression in Eqn~\ref{eqn:product}, we can leverage the close connection between diffusion models and EBMs~\citep{liu2022compositional,du2023reduce}, and learn a set of score functions $\epsilon(\tau, t | l_i)$ for each probability density $p_\theta(\tau|l_i)$. The score of the product of the densities in Equation~\ref{eqn:product} then corresponds to the average of score functions  $\sum_i \frac{1}{N} \epsilon(\tau, t | l_i)$. We can then train this composite score function using the standard denoising diffusion training objective
\begin{equation}
    \label{eqn:comb_denoise}
    \mathcal{L}_{\text{MSE}} = \| \frac{1}{N} \sum_i \epsilon(\tau_t, t | l_i) - \epsilon\|^2,
\end{equation}
where $\tau_t$ corresponds to the original video corrupted with $t$ steps of Gaussian noise.

One issue with directly optimizing Eqn~\ref{eqn:comb_denoise} is that while the product of the composed distribution is encouraged to model the distribution of videos given text $p(\tau|L)$, each component is not necessarily encouraged to accurately model the distribution of videos given relevant textual information in the text snippet $l_i$, $p(\tau|l_i)$. To can encourage the score function $\epsilon(\tau_t, t | l_i)$ to capture this objective by also training the score function to denoise a video given only the relevant text snippet
\begin{equation}
    \label{eqn:single_denoise}
    \mathcal{L}_{\text{MSE}} = \| \epsilon(\tau_t, t | l_i) - \epsilon\|^2.
\end{equation}
To unify both objectives, we use a hybrid training objective where given a set of language components $S = \{l_i \}_{i=1:N}$, we randomly a subset $S'$ of M components and train with objective
\begin{equation}
    \label{eqn:hybrid_denoise}
    \mathcal{L}_{\text{MSE}} = \| \frac{1}{M} \sum_i \epsilon(\tau_t, t | l_{S_i}) - \epsilon\|^2,
\end{equation}
Given these learned score functions, at sampling time, we can sample from a novel combination of score functions. We illustrate overall training and sampling algorithms for our approach in Algorithm~\ref{alg:train} and \ref{alg:inference}.

\begin{algorithm}[tb]
   \caption{Training}
   \label{alg:train}
\begin{algorithmic}[1]
   \STATE {\bfseries Input:} Diffusion Model $\epsilon_\theta$, Training Step N 
   \FOR{$i \ \text{in} \ 0, \dots, N$}
   \STATE Get training samples $\tau_0$ and language instructions $L=\{l_i\}$
   \STATE Diffusion timestep $t \sim \text{Uniform}(\{1, \dots, T\})$
   \STATE $\epsilon \sim \mathcal{N}(\bm{0}, \bm{I})$
   \STATE $\tau_t \leftarrow \sqrt{\bar{\alpha}_t}\tau_0 + \sqrt{1-\bar{\alpha}_t}\epsilon$
   \STATE $\mathcal{L}_{\text{MSE}} =  \|\frac{1}{|L|}\sum_i \epsilon_\theta(\tau_t, t | l_i) - \epsilon\|$
   \STATE Take gradient descent step on $\mathcal{L}_{\text{MSE}}$
   \ENDFOR
\end{algorithmic}
\end{algorithm}
\begin{algorithm}[tb]
   \caption{Inference}
   \label{alg:inference}
\begin{algorithmic}[1]
    \STATE {\bfseries Input:} Diffusion Model $\epsilon_\theta$, Language Instructions $L=\{l_i\}$, Guidance weight $w$
    \STATE $\tau_T \sim \mathcal{N}(\bm{0}, \bm{I})$
    \FOR {$i \ \text{in} \ T, \dots, 1$}
        \STATE $\epsilon_{\text{uncond}} \leftarrow \epsilon_\theta(\tau_t, t)$
        \STATE $\epsilon_i \leftarrow \epsilon_\theta(\tau_t, t|l_i)$
        \STATE $\tilde{\epsilon} \leftarrow = \epsilon_{\text{uncond}} + \sum_i w (\epsilon_\theta(\tau_t, t|l_i) - \epsilon_{\text{uncond}})$
        \STATE $\tau_{t-1} \leftarrow \frac{1}{\sqrt{\alpha_t}} (\tau_t - \frac{1-\alpha_t}{\sqrt{1-\hat{\alpha}_t} \tilde{\epsilon}}) + \sigma_t z$
    \ENDFOR
    \STATE \textbf{Return:} $\tau^0$
\end{algorithmic}
\end{algorithm}

% as long as 
% corresponding image goal specifications $g$, we propose to train our model to compositionally generate.

% To train

% This corresponds to training models to 

% By explicitly training our model to 

% Note that such an approach naturally enables compositional generation -- given a 
% Given a set of sub-instructions and goals, we wish to synthesize an output that is consistent with We propose to train models to compositionally generate outputs. In particular, given 

% We also aim to generate videos under two or more modalities of instructions. Direct combining will cause mutual interference between different modalities and then hurt the performance.
% Previous compositional visual generation works~\cite{} mostly focus on multiple concepts of the text prompt and only in the inference time. We find that it's also essential to learn how to compose. 

% To this end, during training, we propose to 

% The approach to learning how to compose is simple yet effective. We have two stages of training. In the first stage, we only give \method single modality of instruction. In the second stage, we give multi-modal instructions, and the diffusion model will predict noise once for each instruction. The final noise will be the average of all noise.

%sy: Unlike other video diffusion models based on pre-trained text-to-image models, we directly train our models.
% 1. not suitable for robot
% 2. our video generation is based on the first frame. don't need to base on text-to-image

\subsection{Multi-modal Composition}
\label{sect:multimodal}

In addition to conditioning our generation at training time on a set of language components $\{l_i \}_{i=1:N}$ we can also condition our generation of a set of multimodal instructions  $M$ that $M=\{m_i\}_{i=1:K}$. We can express the likelihood of our video generative model $p_\theta(\tau|L, M)$ as the expression 
\begin{equation}
    \label{eqn:product_multimodal}
    \resizebox{0.9\hsize}{!}{$
    p_\theta(\tau|L, M) \propto \prod_{i=1:N} p_\theta(\tau|l_i)^{\frac{1}{N+K}} \prod_{i=1:K} p_\theta(\tau|m_i)^{\frac{1}{N+K}}$},
\end{equation}
corresponding to the product of all models. Note that in the above expression, we can at inference time change the expression to adjust to both a variable number of modalities as well as language, in addition to novel combinations of both modality and language.

Based on the above expression, we can derive a variant of Eqn~\ref{eqn:hybrid_denoise}, to train a set of modality and language conditioned score functions so that they accurately model the above expression in Eqn~\ref{eqn:product_multimodal}
using the objective:
\begin{equation}
    \label{eqn:multi}
    \resizebox{0.9\hsize}{!}{$
    \mathcal{L}_{\text{MSE}} = \|\frac{1}{2M} \sum_i \epsilon(\tau_t, t | l_{S_i}) + \frac{1}{2M} \sum_j \epsilon(\tau_t, t | M_{S_j}) - \epsilon\|^2$},
\end{equation}
Once we have this set of score functions, we can then flexibly compose language and modalities in our compositional world model.
% To encode multimodal information in the form of goal images and goal sketches, we use XXX encoder which we feed into our model.

\begin{figure*}[t]
    \centering
    \includegraphics[width=1.0\linewidth]{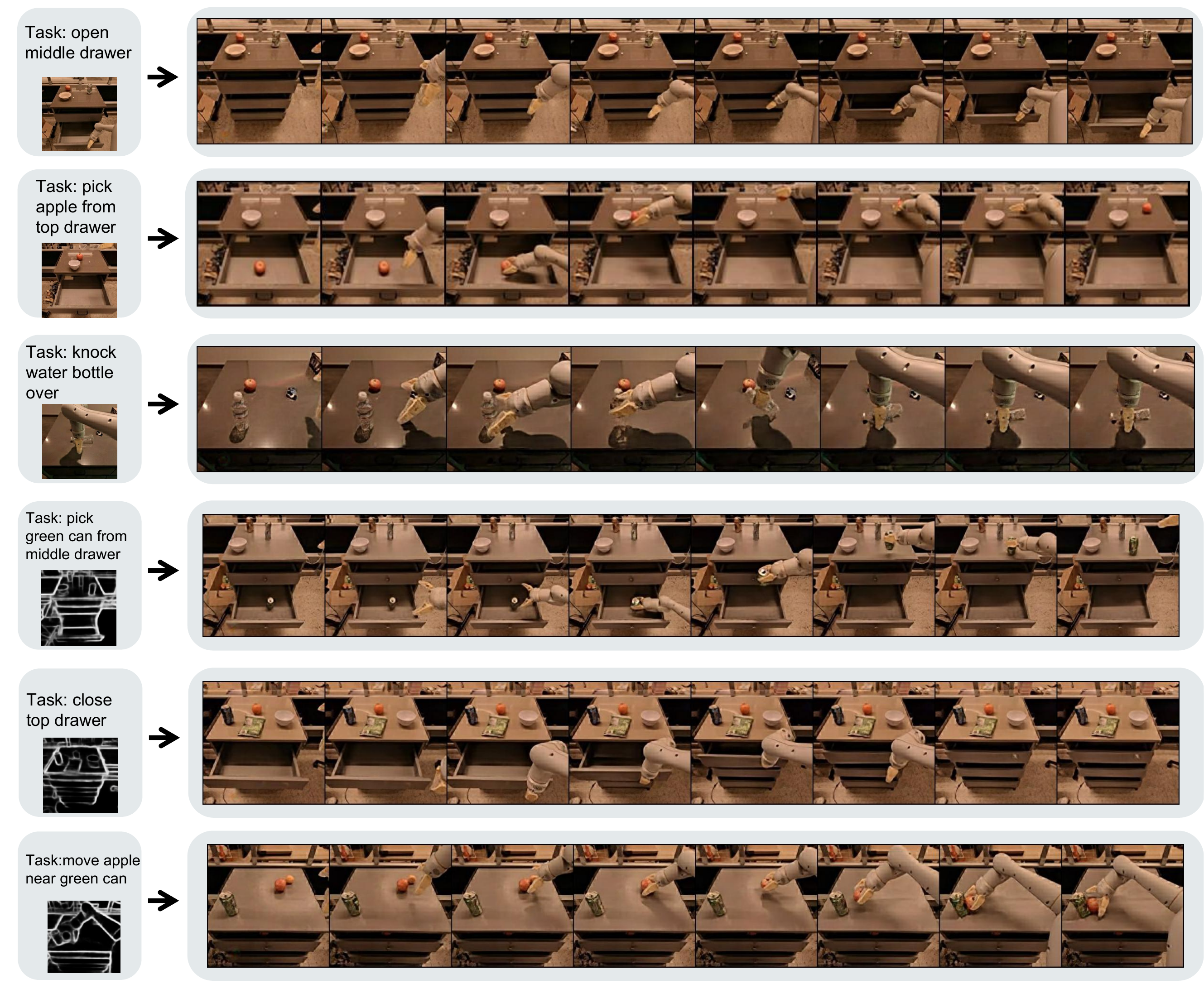}

    \vspace{-3pt}
    \caption{\textbf{Multimodal Compositionality.} \method is able to compose multimodal inputs such as goal and sketch image conditioning with language instructions and synthesize plausible videos.}
    \vspace{-10pt}
    \label{fig:enter-label}
\end{figure*}
% \begin{figure*}
%     \centering
%     \includegraphics{}
%     \caption{\sy{Comparison multi-modal with single-modal and AVDC}}
%     \label{fig:enter-label}
% \end{figure*}

\begin{table}[th]
    \centering
    \begin{tabular}{l c c}
       \toprule
       % & \multicolumn{1}{c}{\bf Seen} & \multicolumn{1}{c}{\bf Unseen} \\
        \textbf{Model}  & \textbf{Seen} &  \textbf{Unseen} \\
        \midrule
        AVDC & 63.1 & 46.9 \\
        HiP & 70.3 & 50.1 \\
        \method w/o & 85.5 & 68.8 \\
        \method & {\bf 90.1} & {\bf 81.3} \\
        \bottomrule
    \end{tabular}
    \caption{{\bf Human Study of Task Instruction Evaluation.} We present human evaluation for object relations and task completion of generated videos on seen and unseen tasks. \method outperforms baselines, especially on unseen tasks.}
    \vspace{-20pt}
    \label{tab:unseen}
\end{table}

\section{Experiments}

In this section, we evaluate the proposed RoboDreamer model in terms of its ability to enable generalizable compositional generation. We structure the experiments to answer the following questions:
\begin{itemize}[leftmargin=*]
    \item \textbf{RQ1:} Does \method have zero-shot generalization abilities when encountering unseen task instructions? (Section \ref{4.1})
    \item \textbf{RQ2:} Does compositional generation of multi-modal instructions improve spatial reasoning and object relations? (Section \ref{4.1})
    \item \textbf{RQ3:} Can \method be deployed on robot manipulation tasks? (Section \ref{4.2})
\end{itemize}

\subsection{Evaluation on Video Generation }\label{4.1}
In this section, we take a comprehensive evaluation of \method's capabilities in two areas: its effectiveness in generating videos from textual instructions through compositional generation and its adeptness in incorporating multi-modal conditions.

\noindent\textbf{Experimental Setup.} We take the real-world robotics dataset RT-1~\cite{brohan2022rt} to evaluate video generation. The dataset consists of various robotic manipulation tasks, i.e. \textit{pick brown chip bag from middle drawer}. The robot is required to detect the middle drawer, pick a brown chip bag, and then place it on the table. Specifically, we train \method on about 70$k$ demonstrations and 500 different tasks. We randomly select language instructions as unseen test cases for evaluation.

\noindent\textbf{Baselines.}  We compare \method with AVDC~\cite{ko2023learning}, a video generation model for robotics;
HiP~\cite{ajay2023compositional}, a latent video diffusion model for robotics;
\method w/o, our model without text-parsing approach.
To make the comparison fair, we only give language instructions to \method and all other baselines.
For all methods, we use pertaining T5-XXL as text encoder.

% \textbf{Dataset.} 

%We have two metrics to evaluate the alignment between video generation and task instructions: human evaluation and detection scores.
\textbf{Metrics.} 
Some works~\cite{girdhar2023emu} have demonstrated that there are still no adequate metrics to evaluate the alignment between video generation and task instructions.
As a result, we conduct human evaluation. Each sample is rated by at least three persons for their task completion. The scores are {0, 1}, where 0 means the robotic planning in the generated videos is unreasonable or fails to solve tasks and 1 means the robotic planning is executable and succeeds in finishing the tasks.
% We use GroundingDINO~\cite{} to evaluate the tasks.
% We also evaluate object spatial relation in the tasks \textit{move A near B} where the robot should pick up the object {\it A} and move to object {\it B} or directly push {\it A} to {\it B}. We use pre-trained open-set object detector GroundingDINO~\cite{liu2023grounding} to detect the object {\it A, B} and measure the distance as detection scores.

\textbf{Implementation Details.} 
The video diffusion model of \method is built upon AVDC and Imagen~\cite{ho2022imagen}.
We use a spatial-temporal convolution network in each ResNet block of U-Net for efficiency. We introduce the temporal attention layer on the ResNet block. We utilize a three-stage cascaded diffusion model for super-resolution.
We use a similar tiling approach to enhance temporal consistency. Since the background of the all frames in one video should be consistent, we concatenate the input condition frame to the all noisy frames before feeding into U-Net.

We use pre-trained models to encode multi-modal instructions.
We use the frozen T5-XXL text encoder~\cite{raffel2020exploring} for processing natural language instructions, which enables us to generate contextual embeddings. 
We borrow the pre-trained image encoder from VQVAE of Stable Diffusion~\cite{rombach2022high} for goal image and goal sketch instructions.
The pre-trained downsampling encoder can be fast to extract spatial information for tasks and enhance efficiency.
All modality embeddings will be fed into PerceiverSampler~\cite{jaegle2021perceiver}, an architecture designed for general inputs and outputs.
The outputs will be integrated into the U-net by introducing a cross-attention layer into the ResNet blocks.

% \subsection{Evaluation on Video Generation }

\noindent\textbf{Zero-shot Generalization.} We first evaluate that the text-parsing approach and compositional generation method bring zero-shot generalization abilities when facing unseen task instructions. The results, as highlighted in Table~\ref{tab:unseen}, reveal a significant enhancement in the model's performance. Through human evaluation, it becomes evident that our approach greatly improves the task alignment of video generation. This underscores the efficacy of \method in understanding and executing unseen task instructions, thereby answering \textbf{RQ1}. As is illustrated in Figure~\ref{fig:zero-shot}, the baseline method AVDC and HiP fail to accurately infer the spatial relationship between objects, incorrectly placing them in the wrong positions. In contrast, our method successfully deduces these relationships, positioning the objects correctly according to the given instructions. By factorizing textual instructions into primitive components, \method could successfully generalize to unseen task instructions by formulating them into combinations of seen components. 

\noindent\textbf{Multi-modal Generation.} Subsequently, we evaluate the multi-modal-conditioned video generation capabilities of \method, focusing on how it leverages visual information to enhance spatial reasoning of video generation. Specifically, we take the final frames as goal images and generate the goal sketches by annotators of ControlNet~\cite{zhang2023adding}. \method's variations were methodically tested: \method (t) is given only language description, \method (t+i) is given language description and the goal image, and \method (t+s) is given language description and goal sketches. As is shown in Table~\ref{tab:multimodal}, with the additional help of goal image instructions or sketch instructions, \method can achieve strong alignment on both human evaluations. This experimental result affirmatively addresses \textbf{RQ2}, with \method's adeptness at integrating multi-modal instructions, \method has a more nuanced understanding and execution of tasks.

\begin{table}[ht]
    \centering
    \begin{tabular}{l c c}
    \toprule
        \textbf{Model}  & \textbf{Human $\uparrow$} & \textbf{FVD $\downarrow$}\\
        \midrule
        AVDC  & 46.9 & 517.1 \\
        \method (t) &  81.3 & 487.8 \\
        \method (t+s) & 94.7 & 454.7 \\
        \method (t+i) & {\bf 95.8} & {\bf 444.3} \\
        \bottomrule
    \end{tabular}
    \caption{{\bf Evaluation on Multi-Modal Generation.} \method (t+s) and \method (t+i) achieve strong performance on human evaluation and good video quality.}
    \vspace{-15pt}
    \label{tab:multimodal}
\end{table}

\begin{table*}[t]
    \centering
    \begin{tabular}{l c c c c c c c}
        \toprule
        \textbf{Model}  & \textbf{lamp off} & \textbf{lamp on} & \textbf{stack blocks} & \textbf{lift block} & \textbf{take shoes} & \textbf{close box} & \textbf{Average} \\
        \midrule
        Image-BC & 60.1 & 47.0 & 0 & 0 & 0 & 82.4 & 31.6 \\ 
        Hiveformer & 81.2 & {\bf 53.2} & 10.6 & {\bf 28.2} & 1.0 & 90.8 & 44.2 \\
        UniPi  & 70.6 & 47.1 & 7.1 & 23.3 & 3.8 & 94.1 & 41.0 \\
        \method & {\bf 96.3} & 51.9 & {\bf 18.5} & 22.2 & {\bf 10.5} & {\bf 96.3} & {\bf 49.3} \\
        \bottomrule
    \end{tabular}
    \caption{{\bf Success Rate on RLBench.} The highest success rate demonstrates that the videos generated by \method are feasible and executable and help robot planning.}
    \vspace{-10pt}
    \label{tab:rlbench}
\end{table*}

\subsection{Evaluation on Robotic Planning} \label{4.2}

Finally, our examination extends to the practical applicability of \method in robotic planning tasks. We investigate whether synthesized videos of \method can do robotic planning.

\noindent\textbf{Experimental Setup.} We conduct experiments on RLBench~\cite{james2020rlbench}. 
The agent captures observations as RGB images using multi-view cameras and controls a robotic arm with seven degrees of freedom ($7$ DoF) on RLBench.
The environment is constructed to mimic real-world conditions, featuring high dimensionality in both observation and action spaces, which presents a significant challenge.
There are 74 challenging vision-based robotic learning tasks whose categories vary from tool-using tasks, and pick-and-place tasks to long-term planning tasks.
We follow the setting of previous works~\cite{guhur2023instruction} to use macro-steps, which will make the environment focus more on robot planning.
We only consider RGB images from the front camera as observations, which makes RLBench much more challenging.
For a fair comparison, we don't add goal images as instructions.

\noindent\textbf{Baselines.} We consider three baselines: 
\begin{itemize}[itemsep=2pt,topsep=0pt,parsep=0pt]
    \item Image-BC, an imitation learning approach given observation, states, and goal description.
    \item Hiveformer~\cite{guhur2023instruction}: a transformer-based approach that integrates natural language instructions, multi-view scene observations, and a full history of observations and actions.
    \item UniPi~\cite{du2023learning}: a method that utilizes text-to-video models to generate videos and inverse dynamic policy to predict actions based on the videos. We use the open-source text-to-video codebase from ~\citep{ko2023learning} to train models.
\end{itemize}

\noindent\textbf{Robot Planning.} According to the results presented in Table~\ref{tab:rlbench}, \method achieves superior task success rates compared to baseline models even if \method is only given observation from single cameras. 
As expected, Image-BC and Hiveformer are struggling in the long-term tasks that are \textit{stack blocks} and \textit{take shoes}. On the other hand, \method achieves a success rate of 15\% with the help of predicted future observations.
UniPi performs poorly as it does not align with task instructions well.
The strong performance of \method demonstrates that the synthesized videos are significantly beneficial for robot planning.

% This performance conclusively answers our \textbf{RQ3}. 

% \subsection{Discussions}

\section{Related Work}

% diffusion models for decision-making
% compositional generation

% \textbf{Video Generations}

\textbf{Diffusion Models for Decision-Making} Diffusion models have emerged as promising generative models for many decision-making tasks~\cite{chi2023diffusion,pearce2023imitating,zhang2022lad,liang2023adaptdiffuser,huang2023diffusion,structdiffusion2023,zhou2023adaptive}. Some works~\cite{janner2022planning,ajay2022conditional} train diffusion model on low-level state and action space on simulation data. They generate trajectories to do robot planning.
While these works are hard to generalize to high-dimensional data like videos, some works~\cite{du2023learning,ko2023learning} formulated the robot planning as a text-to-video generation problem. Most similar to our work, UniPi~\cite{du2023learning} trains a video diffusion model to predict future frames and gets actions with inverse dynamic policy. On the other hand, AVDC~\cite{ko2023learning} utilizes pre-trained flow networks to predict the actions. However, previous works are limited in generalization to unseen tasks. Our text-parsing approach and composition abilities can enhance the zero-shot generalization abilities of \method.

\textbf{Compositional Generation}
Our work is also related to the compositional generation method~\cite{bar2020compositional,liu2022compositional, yu2022modular,ajay2023compositional, yang2023probabilistic,zhang2023adding,shi2023exploring,hu2024instruct,du2023reduce}. The promising solutions are based on those works~\cite{nie2021controllable,du2021unsupervised,liu2021learning,gkanatsios2023energy} that probabilistically combing different generative models for jointly generating outputs. Most of them works have delved into the realm of compositional text~\cite{deng2020residual,liu-etal-2023-composable} and image generation~\cite{shi2023exploring,liu2022compositional,du2023reduce}. However, when it comes to the domain of text-to-video generation, the extent of exploration of compositionality is  comparatively limited. In text-to-video generation, VideoAdapter~\cite{yang2023probabilistic} introduces a novel setting by transferring pre-trained general T2V models to domain-specific T2V tasks with small amount of training data, however, it fails to genelize on unseen tasks or text descriptions. HIP~\cite{ajay2023compositional} aims to solve long-horizon tasks, however, it composes several expert foundation models. Different from previous research, our approach demonstrates how to construct compositional video-based world models,  decomposing the learned probability distribution during training into a series of compositional components. This empowers us to seamlessly generate videos subject to combinations of detailed language and image specifications that are not seen during training. 
% \section{Discussion and Limitations}
% \vspace{-5pt}
\section{Conclusion}

In conclusion, we introduce \method, a compositional approach to video generation that generalizes significantly better than prior works in video generation in robotics. By leveraging the natural compositionality of language and integrating multi-modal instructions, \method demonstrates significant advancements in generating videos that accurately capture complex spatial relationships and object interactions. Experimental results verify \method's capabilities in zero-shot generalization, multi-modal-conditioned video generation, and its potential application in robotic manipulation tasks.

\looseness=-1
\textbf{Limitations} Although \method exhibits strong performance in robot planning tasks, it has several limitations. \textbf{(1)} While many robot tasks often use information from multiple cameras, \method is limited to single camera views and is unable to consider the multi-camera information. This limits the applicability of \method to many robotics tasks that require detailed 3D information. We believe that exploring how to add 3D inductive biases into \method to consider multi-camera information is a rich source of future research. \textbf{(2)} \method generalizes poorly to many real-world images we tested. We believe that existing robotics datasets are still limited in diversity, and it may interesting to explore joint co-training of our compositional model across both robotics data and existing videos on YouTube to improve generalization. \textbf{(3)} The capabilities of video generation models, including ours, are constrained when it comes to moving-camera settings. Addressing these challenges necessitates the approach to stabilizing.

% \textbf{Conclusion}

\section*{Impact Statement}

This paper presents work whose goal is to advance the field of Machine Learning. There are many potential societal consequences of our work, none of which we feel must be specifically highlighted here. 


\bibliography{icml2024}
\bibliographystyle{icml2024}

%%%%%%%%%%%%%%%%%%%%%%%%%%%%%%%%%%%%%%%%%%%%%%%%%%%%%%%%%%%%%%%%%%%%%%%%%%%%%%%
%%%%%%%%%%%%%%%%%%%%%%%%%%%%%%%%%%%%%%%%%%%%%%%%%%%%%%%%%%%%%%%%%%%%%%%%%%%%%%%
% APPENDIX
%%%%%%%%%%%%%%%%%%%%%%%%%%%%%%%%%%%%%%%%%%%%%%%%%%%%%%%%%%%%%%%%%%%%%%%%%%%%%%%
%%%%%%%%%%%%%%%%%%%%%%%%%%%%%%%%%%%%%%%%%%%%%%%%%%%%%%%%%%%%%%%%%%%%%%%%%%%%%%%
\appendix
\onecolumn

\section{Experimental Details}

\subsection{Video Diffusion}

We list the details about our method \method as follows.
\begin{enumerate}[align=right,itemindent=0em,labelsep=2pt,labelwidth=1em,leftmargin=*,itemsep=1pt]
\item Our method \method is built upon AVDC~\cite{ko2023learning} and Imagen~\cite{ho2022imagen} and we utilize a three-stage cascaded diffusion model for super-resolution.
\item For video diffusion models, we use 4 ResNetBlock within U-Net and each block is composed of spatial-temporal convolution layers and cross-attention layers with conditioned instructions.
\item We introduce temporal-attention layers in the last block within the encoder of U-Net and the first block within the decoder.
\item The base channel is 128 and the channel multiplier is [1, 2, 4, 8].
\item We train our video diffusion models with 256 batch size and 5e-5 learning rate on about 100 V100 GPUs.
\item We train base video diffusion models with $8 \times 64 \times 64$ videos and then subsequently upsample to $8 \times 128 \times 128$ and $8 \times 256 \times 256$ videos.
\end{enumerate}

\subsection{Other Details}

\begin{enumerate}[align=right,itemindent=0em,labelsep=2pt,labelwidth=1em,leftmargin=*,itemsep=1pt]

\item RT-1 Dataset: RT-1 Dataset has about $70k$ demonstrations with an average length of 44. We sample one every 5 frames. There are about 500 tasks. We list some categories of tasks here: \textit{pick}, \textit{pick ... from ...}, \textit{place}, \textit{open}, \textit{close}, \textit{knock} and \textit{pull}. 
\item Inverse Dynamics Model: Inverse dynamics model is trained to predict actions given two adjacent frames and the current state. We use ResNet18 as the backbone followed by an MLP layer. This model is trained using Adam optimizer with a learning rate 1e-4 for 10K steps. 
\item  RLBench: We use Franka Panda arm with Franka gripper on RLBench. It has seven degrees of freedom (7 DoF) and 8-dimensional action space, along with an additional gripper state.
\item Human Evaluation: Each sample is rated by at least three human raters and we evaluate about 128 samples in total including more than 20 text prompts.

\end{enumerate} 

\section{Additional Results}

% \subsection{Text Parsing Results}

% We show our text parsing results of language instructions from different datasets in Table~\ref{tab:parse_example}.

% \input{tables/parse_example}

\subsection{Visualization on RLBench}

\begin{figure}[ht]
    \centering
    \includegraphics[width=1.0\linewidth]{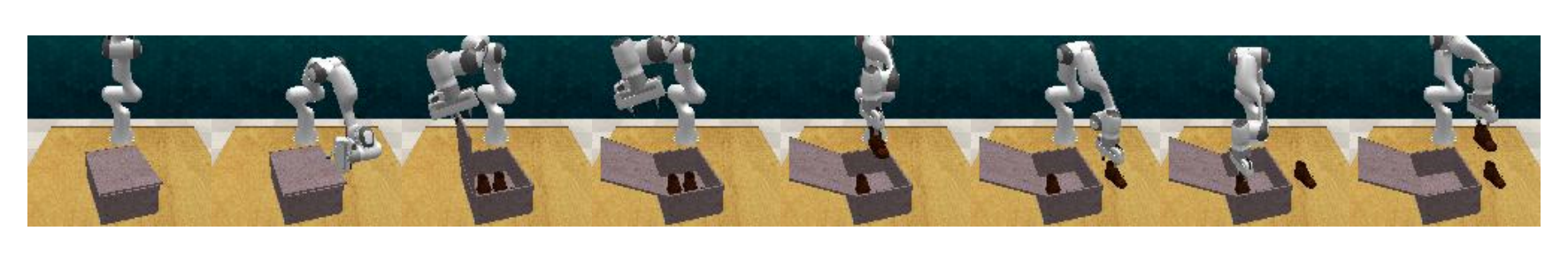}
    \caption{{\bf Visualization on RLBench.}}
    \label{fig:rlb_vis}
\end{figure}

We visualize the tasks on RLBench in Figure~\ref{fig:rlb_vis}.

\subsection{More Results on Video Generation}

More video generation results are shown in the website (https://robovideo.github.io/).

\begin{figure}[t]
    \centering
    \includegraphics[width=1.0\linewidth]{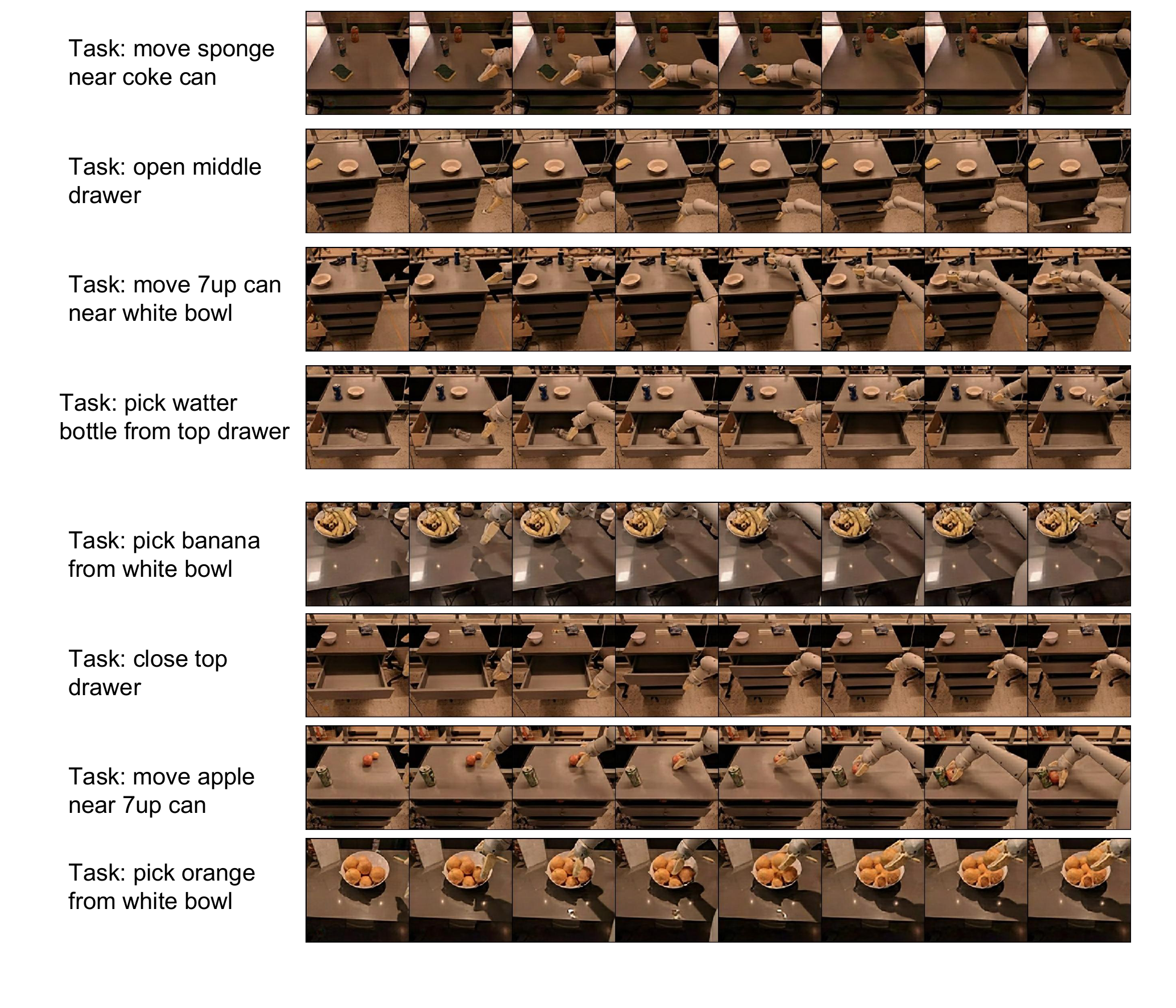}
    \vspace{-30pt}
    \caption{{\bf Video Generation.}}
    \label{fig:rlb_vis123}
\end{figure}

\subsection{IMO Metrics}

We use pretrained GroundingDino to detect the bounding boxes of the target objects. The IMO results (Iou of the bounding boxes) are shown in Table~\ref{tab:imo}. The results are consistent with Human Evaluation. This supports the effectiveness of our approach in achieving better alignment through multi-modal inputs.

\begin{table}[ht]
    \centering
    \vspace{-80pt}
    \begin{tabular}{l c c c}
    \toprule
        \textbf{Model}  & \textbf{Human $\uparrow$} & \textbf{FVD $\downarrow$} & \textbf{IMO$\uparrow$} \\
        \midrule
        \method (t) &  81.3 & 487.8 & 63.5 \\
        \method (t+s) & 94.7 & 454.7 & 72.5 \\
        \method (t+i) & {\bf 95.8} & {\bf 444.3} & {\bf 78.1} \\
        \bottomrule
    \end{tabular}
    \caption{{\bf Evaluation on Multi-Modal Generation.} \method (t+s) and \method (t+i) achieve strong performance on human evaluation and good video quality.}
    \vspace{-15pt}
    \label{tab:imo}
\end{table}

% You can have as much text here as you want. The main body must be at most $8$ pages long.
% For the final version, one more page can be added.
% If you want, you can use an appendix like this one.  

% The $\mathtt{\backslash onecolumn}$ command above can be kept in place if you prefer a one-column appendix, or can be removed if you prefer a two-column appendix.  Apart from this possible change, the style (font size, spacing, margins, page numbering, etc.) should be kept the same as the main body.
%%%%%%%%%%%%%%%%%%%%%%%%%%%%%%%%%%%%%%%%%%%%%%%%%%%%%%%%%%%%%%%%%%%%%%%%%%%%%%%
%%%%%%%%%%%%%%%%%%%%%%%%%%%%%%%%%%%%%%%%%%%%%%%%%%%%%%%%%%%%%%%%%%%%%%%%%%%%%%%

\end{document}